\documentclass[10pt,twocolumn,a4paper]{article}

\usepackage[utf8]{inputenc}
\usepackage{amsmath,amssymb,amsfonts}
\usepackage{graphicx}
\usepackage{booktabs}
\usepackage{tabularx}
\usepackage{multirow}
\usepackage{subcaption}
\usepackage{microtype}
\usepackage{cite}
\usepackage{float}
\usepackage{url}
\usepackage{authblk}
\usepackage[colorlinks=true, linkcolor=blue, citecolor=blue, urlcolor=blue]{hyperref}

\usepackage{arxiv}

\title{\LARGE\textbf{Real-Time Conveyor Belt Defect Detection via Edge TPU-Accelerated YOLOv11 on Raspberry Pi 5}}

\author{Muhammad Nur Bijak Bestari\textsuperscript{1,*} \quad Mulki Indana Zulfa\textsuperscript{1}}

\date{}

\begin{document}

\twocolumn[
  \begin{@twocolumnfalse}
    \maketitle
    \vspace{-2.5em}
    \begin{center}
      {\small \textsuperscript{1}Department of Electrical Engineering, Faculty of Engineering, Universitas Jenderal Soedirman, Purwokerto 53123, Central Java, Indonesia}
    \end{center}
    \vspace{0.8em}

    \begin{abstract}
    \noindent Conveyor belts are vital operational arteries in continuous bulk material handling, particularly in heavy process industries such as cement manufacturing and mining. Undetected structural anomalies---including longitudinal tears, localized holes, degraded patch repairs, and impact punctures---can lead to catastrophic belt snapping, costly unplanned production downtime, and severe occupational safety hazards. Conventional manual visual inspection is inherently periodic, labor-intensive, hazardous, and subjective. Although deep learning vision systems offer high detection accuracy, existing models are predominantly evaluated on power-intensive workstation GPUs or cloud infrastructures, hindering practical deployment at remote, resource-constrained industrial edge points. In this paper, we propose, implement, and empirically validate an end-to-end, low-power edge AI vision system for real-time conveyor belt damage detection deployed on a Raspberry Pi 5 single-board computer accelerated by a Google Coral Edge TPU coprocessor. The system leverages an optimized YOLOv11 nano (YOLOv11n) architecture trained on a curated multi-source industrial corpus exceeding 3,000 augmented samples, integrating in-situ plant inspection imagery from a prominent cement production facility in West Java, open-source industrial datasets, and autonomous time-lapse captures. To achieve real-time throughput on edge hardware, the model undergoes post-training full integer quantization (INT8) and is compiled specifically for the Edge TPU systolic tensor architecture. Experimental results demonstrate solid detection capability, achieving an overall precision of 0.742, recall of 0.730, mAP@0.5 of 0.764, and mAP@0.5:0.95 of 0.553 across four defect classes. Deployed on-site in an enclosed industrial panel box and interfaced with an RTSP-enabled industrial IP camera over Power over Ethernet (PoE), the system achieves stable inference throughput of 30--40 frames per second (FPS) at $320 \times 320$ resolution with an average end-to-end latency of 70--90 ms. Long-term operational profiling under harsh clinker dust and mechanical vibration confirms thermal equilibrium at $55 \pm 5^\circ\text{C}$ on the host CPU and $\sim 65^\circ\text{C}$ on the Edge TPU with zero frame loss. The proposed framework demonstrates that edge AI can bridge the gap between high-accuracy computer vision and practical, cost-effective predictive maintenance in continuous manufacturing operations.
    \end{abstract}

    \vspace{0.5em}

    \vspace{1.2em}
  \end{@twocolumnfalse}
]

{
  \renewcommand{\thefootnote}{\fnsymbol{footnote}}
  \footnotetext[1]{Corresponding author. E-mail: \texttt{muhammad.bestari@mhs.unsoed.ac.id}, \texttt{mbijakbestari@gmail.com}.}
}

\section{Introduction}
\label{sec:intro}

Continuous material transportation systems constitute the critical infrastructural backbone of modern heavy manufacturing, mineral extraction, and cement production facilities~\cite{rzeszowska2025analysis}. Among these systems, heavy-duty rubber conveyor belts operate uninterruptedly over hundreds of meters, conveying thousands of tons of abrasive raw materials, crushed limestone, hot clinker, and finished additives daily. Given the aggressive operating environment characterized by high mechanical tension, impact from dropped rock fragments, abrasive friction, and elevated temperatures, conveyor belts are constantly susceptible to surface degradation and catastrophic structural failure~\cite{guo2022belt}. The most prevalent defect topologies include longitudinal tears, localized puncture wounds, through-thickness holes, and peeling or uneven patch repairs (patch work)~\cite{liu2025yolo}. If minor tears or punctures remain undetected during normal operation, shear forces and dynamic pulley tension can rapidly propagate the defect into full longitudinal belt rip, resulting in prolonged production outages, substantial raw material spillage, and repair expenses exceeding tens of thousands of dollars per incident~\cite{hu2025real}.

Historically, condition assessment of industrial conveyor belts has relied heavily on periodic manual walk-through inspections conducted by plant maintenance personnel. However, manual inspection exhibits profound limitations: it is labor-intensive, hazardous to personnel working in close proximity to high-speed rotating idlers, subjective to operator alertness, and incapable of continuous real-time coverage~\cite{guo2023machine}. Minor cracks and internal cord tears frequently escape visual notice until catastrophic failure occurs. Furthermore, time-based scheduled maintenance strategies often fail to predict sudden mechanical failures resulting from unpredicted foreign body intrusion, introducing substantial operational uncertainty into plant maintenance planning~\cite{rzeszowska2025analysis}.

To overcome the vulnerabilities of manual inspection, non-destructive testing (NDT) technologies---such as X-ray radiography, electromagnetic induction coil sensors, and ultrasonic scanning---have been explored~\cite{guo2022belt}. Although specialized NDT systems are capable of inspecting internal steel cord integrity, they involve prohibitively high procurement costs, complex physical installations around conveyor frames, and radiation safety considerations. Moreover, standard electromagnetic and ultrasonic detectors struggle to reliably identify superficial elastomeric damage, such as delaminating vulcanized rubber patches or shallow surface tears, which are precursors to complete belt failure.

Recent advancements in computer vision and deep learning have catalyzed the adoption of digital image processing for automated industrial inspection~\cite{guo2023machine}. Single-stage object detectors, particularly the You Only Look Once (YOLO) family, have demonstrated remarkable success in industrial inspection benchmarks owing to their favorable trade-off between detection accuracy and inference latency~\cite{liu2025yolo,hu2025real}. For example, Liu et al.~\cite{liu2025yolo} introduced YOLO-STOD, an attention-augmented YOLOv5 variant designed for conveyor tear detection in coal mines, while Hu et al.~\cite{hu2025real} developed a lightweight YOLO architecture to improve small-defect sensitivity. However, an analysis of the existing literature reveals a critical \textbf{research gap}: the vast majority of prior studies validate their models exclusively on high-performance workstation GPUs (e.g., NVIDIA RTX 3080/4090 or cloud clusters) using static offline datasets. Such high-power configurations fail to address the stringent operational realities of industrial edge deployment---namely, strict power limits, harsh dust and thermal constraints, limited embedded RAM, and the latency/bandwidth penalties associated with transmitting raw video streams over industrial field networks to centralized servers~\cite{prokscha2023efficient,chen2022rewriting}.

To address this deployment gap, this paper presents the design, embedded optimization, and in-situ industrial validation of an autonomous, low-power edge AI vision system for real-time conveyor belt defect detection. The core contributions of this work are fourfold:
\begin{enumerate}
    \item \textbf{Edge-Optimized YOLOv11 Architecture:} We adapt and optimize the recent YOLOv11 nano (YOLOv11n) architecture for industrial edge hardware, executing post-training full integer quantization (INT8) to map 100\% of operations directly onto the systolic tensor cores of the Google Coral Edge TPU.
    \item \textbf{Curated Multi-Source Industrial Corpus:} We construct and curate an extensive dataset comprising over 3,000 augmented images derived from real industrial field inspections at a major cement manufacturing facility in West Java, converted open-source polygon datasets, and autonomous time-lapse captures.
    \item \textbf{Comprehensive Execution Benchmarking:} We conduct rigorous comparative latency and throughput evaluations across PyTorch FP32 CPU, ONNX Runtime CPU, and Edge TPU INT8 across multiple spatial resolutions ($280\times 280$, $320\times 320$, and $640\times 640$ pixels).
    \item \textbf{Empirical Field Validation \& Thermal Profiling:} We validate the complete end-to-end pipeline---interfacing directly with an on-site industrial RTSP IP camera over PoE inside an IP-rated industrial enclosure---quantifying real-time FPS, end-to-end pipeline latency, and thermal equilibrium under harsh industrial plant conditions.
\end{enumerate}

The remainder of this article is organized as follows. Section~\ref{sec:related} reviews related work in conveyor monitoring and edge inference. Section~\ref{sec:method} details the system architecture, dataset curation, model training, and Edge TPU compilation pipeline. Section~\ref{sec:results} reports the experimental results, benchmark comparisons, and field evaluation findings. Section~\ref{sec:discussion} discusses engineering implications, trade-offs, and operational limitations. Finally, Section~\ref{sec:conclusion} concludes the paper with future research avenues.

\section{Related Work}
\label{sec:related}

\subsection{Conveyor Belt Condition Monitoring and Defect Inspection}
Condition assessment of industrial belt conveyors has evolved from rudimentary manual observation to automated instrumentation. Rzeszowska et al.~\cite{rzeszowska2025analysis} demonstrated that conventional time-based maintenance indicators suffer from significant uncertainty due to non-uniform operational wear, varying tonnage loading, and unpredictable foreign object intrusion. Consequently, visual monitoring has emerged as an indispensable modality for real-time defect localization. Guo et al.~\cite{guo2022belt} conducted a comprehensive survey comparing optical vision, X-ray scanning, and inductive sensors for coal conveyor inspection, highlighting that while X-ray can inspect internal steel cords, optical vision offers unmatched flexibility, non-hazardous operation, and lower capital expenditure. Guo et al.~\cite{guo2023machine} subsequently proposed a machine vision framework utilizing fusion knowledge distillation to improve surface damage segmentation under uneven lighting conditions.

\subsection{Deep Learning Detectors in Industrial Applications}
The YOLO architecture family has achieved widespread adoption in manufacturing inspection due to its unified, single-stage regression formulation~\cite{zhang2024improved}, with recent models continuously pushing end-to-end real-time efficiency~\cite{jocher2026yolo26}. In the domain of conveyor belt safety, Liu et al.~\cite{liu2025yolo} proposed YOLO-STOD, incorporating coordinate attention and a small-object detection head into YOLOv5, achieving improved precision on coal mine tear samples. Similarly, Hu et al.~\cite{hu2025real} introduced structural modifications to YOLO to balance speed and accuracy. However, both studies performed evaluation exclusively on desktop-class GPUs without demonstrating embedded compatibility or analyzing quantization loss on integer-only edge coprocessors.

\subsection{Edge AI and Hardware Acceleration via Edge TPU}
Deploying deep convolutional neural networks onto resource-constrained embedded platforms requires aggressive optimization, including weight pruning and post-training quantization~\cite{chen2022rewriting,reddi2020mlperf}. Google's Edge TPU is an application-specific integrated circuit (ASIC) designed specifically for executing full 8-bit integer (INT8) neural network operations at high power efficiency ($\sim 2\text{ TOPS/W}$, delivering up to 4 TOPS peak compute)~\cite{coral2023edgetpu}. Prokscha et al.~\cite{prokscha2023efficient} demonstrated that deploying YOLOv5 onto a Google Coral Edge TPU dramatically accelerates inference compared to embedded ARM CPUs, reducing power draw to under 5 W. Ghosh et al.~\cite{ghosh2020assistive} validated the combination of Raspberry Pi and Coral USB Accelerator for real-time mobile assistance. Nevertheless, compiling the latest YOLO architectures (such as YOLOv11) for the Edge TPU involves strict architectural constraints: every operator in the inference graph must be quantized to INT8 with static tensor shapes, and non-supported tensor operations fall back to the host CPU, causing severe latency spikes~\cite{chen2022rewriting}. This work provides the first systematic study of compiling, deploying, and profiling YOLOv11 on the Raspberry Pi 5 and Google Coral Edge TPU pipeline in an active industrial manufacturing plant.

\section{System Methodology and Implementation}
\label{sec:method}

\subsection{Overall Research Flow}
\begin{figure}[!htbp]
    \centering
    \includegraphics[width=\linewidth]{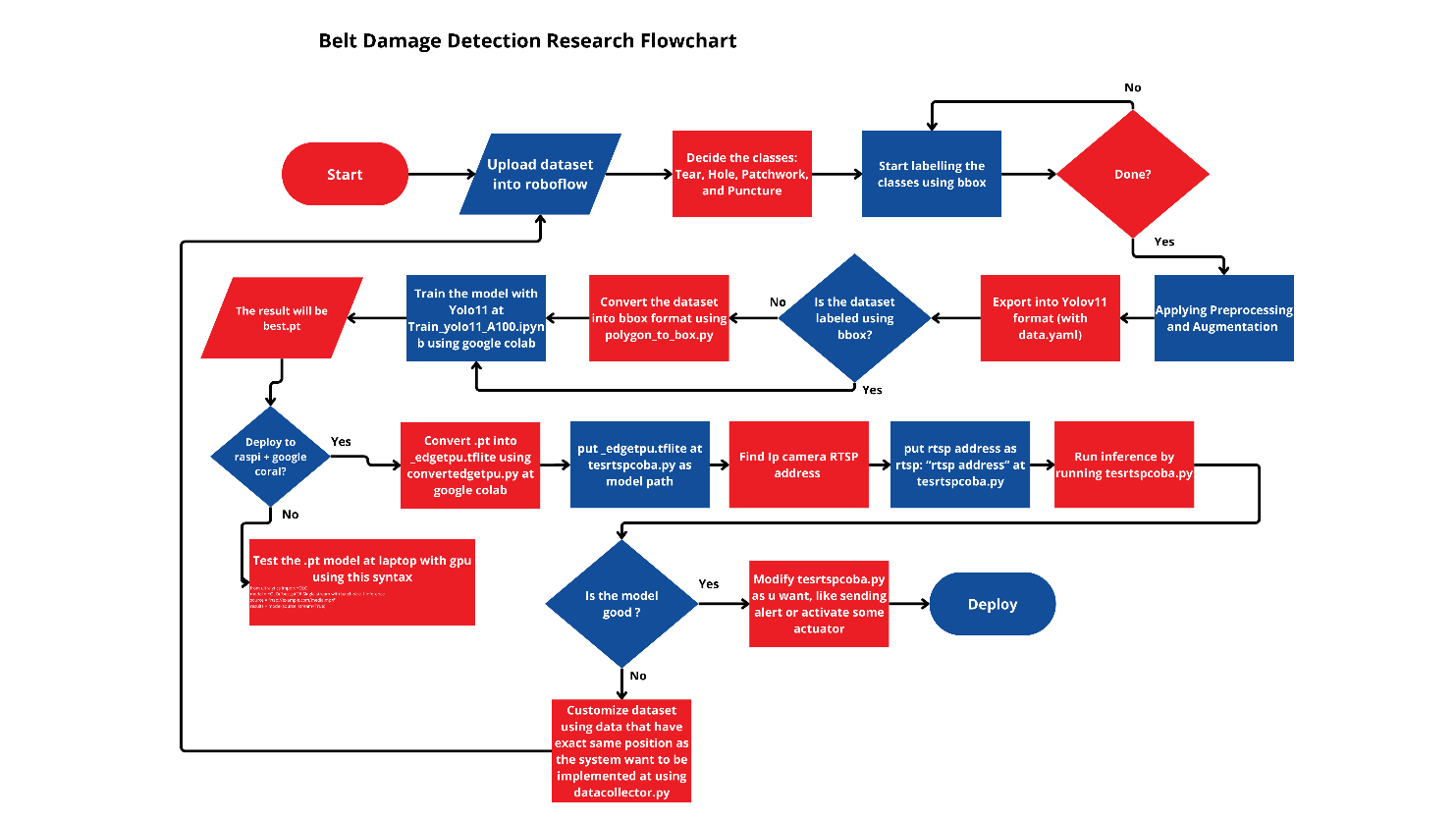}
    \caption{End-to-end research methodology flowchart, delineating data acquisition, annotation conversion, YOLOv11 model training, INT8 Edge TPU compilation, and embedded industrial RTSP deployment.}
    \label{fig:flowchart}
\end{figure}

The methodological workflow adopted in this research is illustrated in Fig.~\ref{fig:flowchart}. It comprises four principal stages: (1) multi-source dataset collection, annotation, and augmentation; (2) baseline model training and validation using the YOLOv11 architecture on high-performance cloud compute; (3) model conversion, full integer INT8 quantization, and hardware compilation via the Edge TPU Compiler; and (4) embedded deployment, RTSP video pipeline integration, and industrial field testing on a Raspberry Pi 5 platform.

\subsection{System Architecture and Industrial Hardware Design}
\begin{figure}[!htbp]
    \centering
    \includegraphics[width=\linewidth]{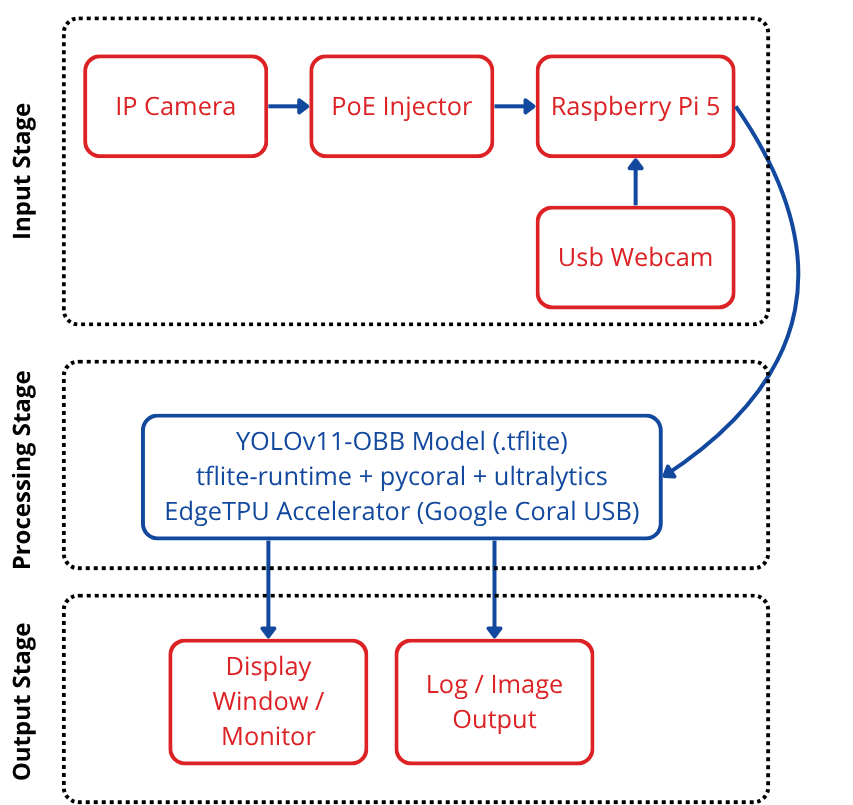}
    \caption{Three-tier edge computing system architecture, depicting the RTSP IP camera feed, PoE injector, Raspberry Pi 5 host, Google Coral Edge TPU coprocessor, and real-time monitoring display.}
    \label{fig:arch}
\end{figure}

The end-to-end hardware architecture is structured into three functional tiers, as depicted in Fig.~\ref{fig:arch}:
\begin{enumerate}
    \item \textbf{Data Acquisition Tier:} An industrial-grade Hikvision DS-2CD2T26G2-4I IP camera is deployed above the conveyor belt transfer point. The camera features a 2 MP sensor ($1920 \times 1080$ resolution), H.265+/H.264 hardware encoding, DarkFighter ultra-low-light capability, and 120 dB True Wide Dynamic Range (WDR), which mitigates high-contrast shadows created by industrial overhead lighting. Power and bidirectional high-speed data transmission are provided over a single shielded Cat6 cable via a TP-Link POE150S Gigabit PoE Injector (48 V DC, 15.4 W).
    \item \textbf{Edge Processing Tier:} The processing core is a Raspberry Pi 5 embedded single-board computer powered by a Broadcom BCM2712 Quad-core ARM Cortex-A76 processor clocked at 2.4 GHz with 16 GB of high-speed LPDDR4X-4267 SDRAM, running 64-bit Raspberry Pi OS (Debian 12 Bookworm). Hardware-accelerated inference is offloaded to a Google Coral USB Accelerator containing an Edge TPU ASIC (4 TOPS peak compute at INT8 precision) connected over a dedicated USB 3.0 interface.
    \item \textbf{Supervisory and Output Tier:} An industrial dashboard renders real-time bounding box predictions, class labels, and confidence probabilities on an operator display. The host application continuously streams the processed video, stores alert logs, and interfaces with plant management protocols.
\end{enumerate}

\begin{figure}[!htbp]
    \centering
    \begin{subfigure}[b]{0.48\linewidth}
        \centering
        \includegraphics[width=\linewidth]{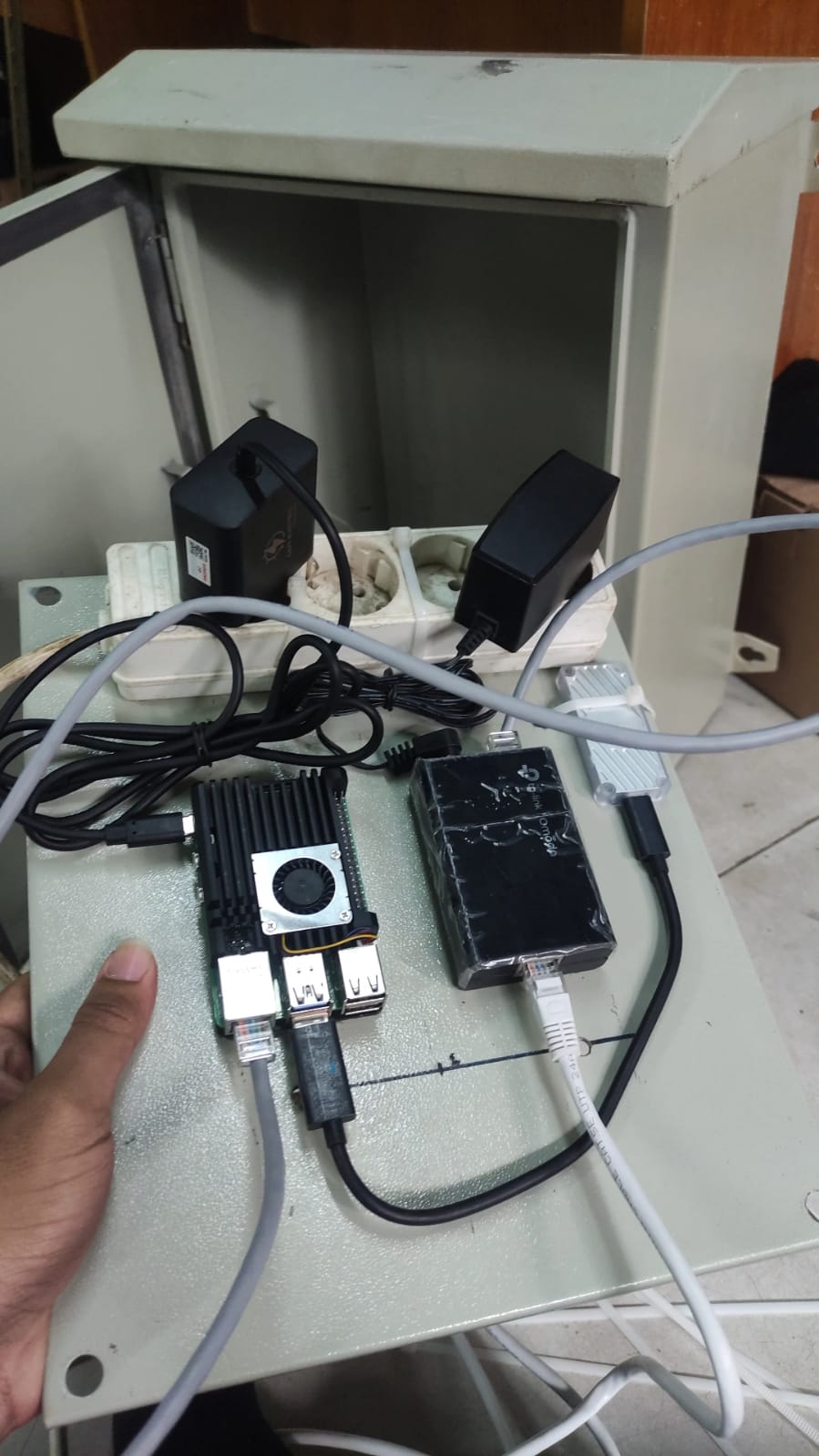}
        \caption{Sealed panel enclosure}
    \end{subfigure}
    \hfill
    \begin{subfigure}[b]{0.48\linewidth}
        \centering
        \includegraphics[width=\linewidth]{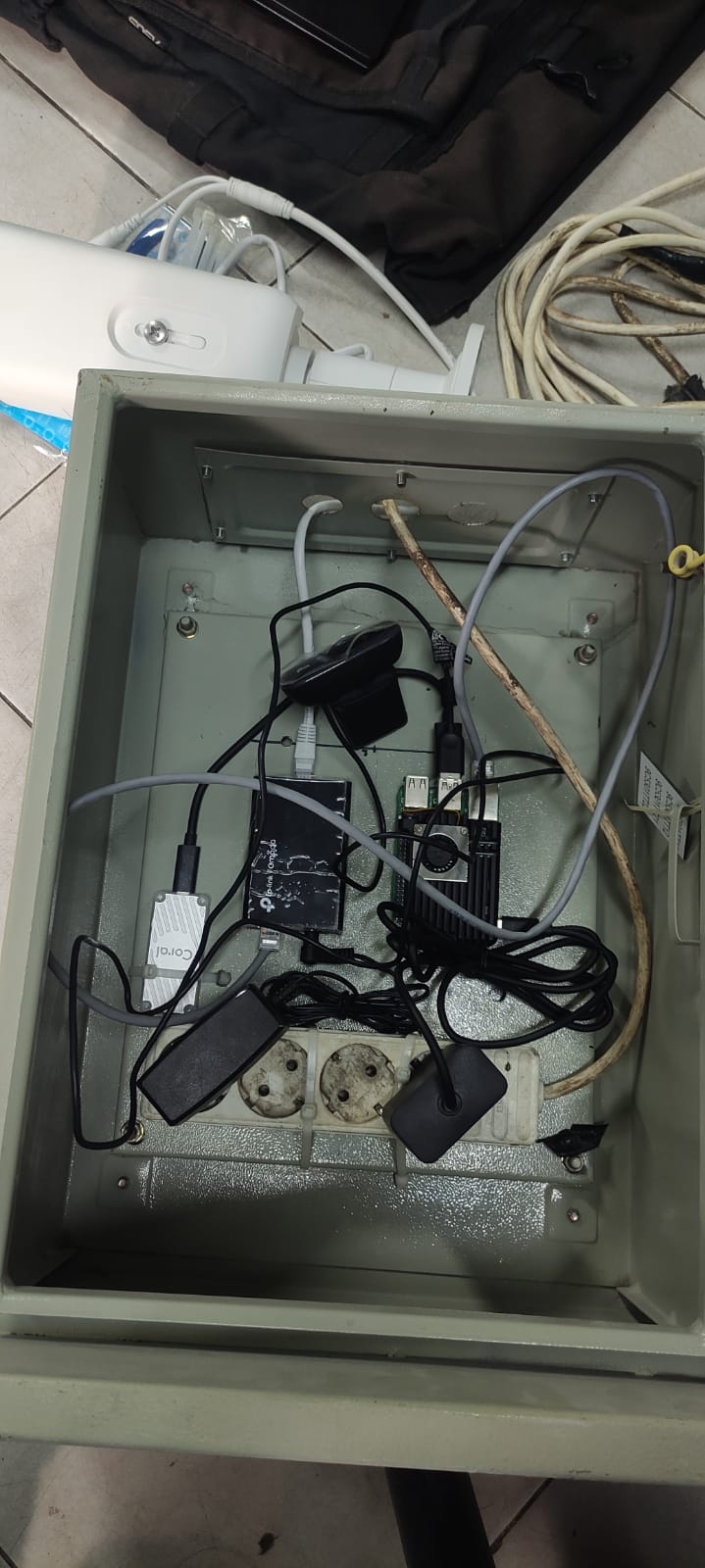}
        \caption{Laboratory test bench}
    \end{subfigure}
    \caption{Industrial hardware deployment: (a) on-site IP-rated protective metal enclosure containing the Raspberry Pi 5, Edge TPU, and PoE supply at the cement facility; (b) laboratory hardware testing bench.}
    \label{fig:hardware_deployment}
\end{figure}

To ensure long-term survivability in the cement plant environment, all edge processing hardware is housed inside a heavy-duty metallic IP-sealed industrial enclosure box (Fig.~\ref{fig:hardware_deployment}), shielding the sensitive electronics from ambient cement clinker dust, ambient moisture, and mechanical conveyor vibrations.

\subsection{Multi-Source Industrial Dataset Curation}
Supervised training of robust deep learning models in industrial vision requires diverse training distributions that reflect actual plant illumination, material dust accumulation, and surface texture irregularities~\cite{murrugarra2022can}. To achieve this, we synthesized a comprehensive dataset from three complementary sources:
\begin{enumerate}
    \item \textbf{In-Situ Field Inspection Data:} Captured directly at the production facilities of an industrial cement manufacturing plant in West Java, Indonesia. This dataset includes 8 high-resolution video recordings of operational conveyor belts exhibiting longitudinal tears, 99 field photographs of tears, and 35 photographs of vulcanized patch repairs (patch work) under varying daylight and artificial plant illumination. Video files were decomposed into static frames using a dedicated frame extraction script.
    \item \textbf{Public Defect Corpus with Annotation Transformation:} A public dataset comprising 922 images of conveyor belt damage from Roboflow Universe was integrated. The raw annotations were provided as polygon segmentation masks across four classes: Tear (689 instances), Patch Work (405 instances), Puncture (120 instances), and Hole (42 instances). Because the Edge TPU compiler does not support dense instance segmentation heads with dynamic post-processing operators, we developed a specialized automated conversion script (\texttt{polygon\_to\_box.py}) to transform all arbitrary polygon masks into precise oriented and axis-aligned bounding boxes (OBB / BBox), ensuring 100\% compatibility with Edge TPU integer detection runtimes.
    \item \textbf{Autonomous Time-Lapse Collector:} An automated software utility (\texttt{data\_collector.py}) was deployed on the edge camera node to record 5--10 minute intervals of continuous belt movement, extracting frames periodically according to pre-configured time steps while enforcing a Region of Interest (ROI) mask to exclude static peripheral structures.
\end{enumerate}

\begin{figure}[!htbp]
    \centering
    \includegraphics[width=\linewidth]{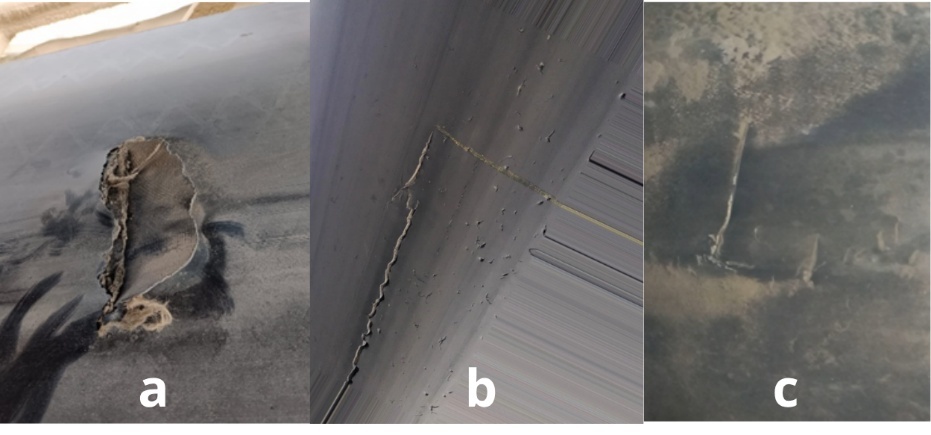}
    \caption{Representative raw conveyor belt samples: (a) direct in-situ field inspection captures at the cement plant; (b) public industrial dataset samples; (c) automated frames from the stream sampler.}
    \label{fig:raw_samples}
\end{figure}

\begin{figure}[!htbp]
    \centering
    \includegraphics[width=\linewidth]{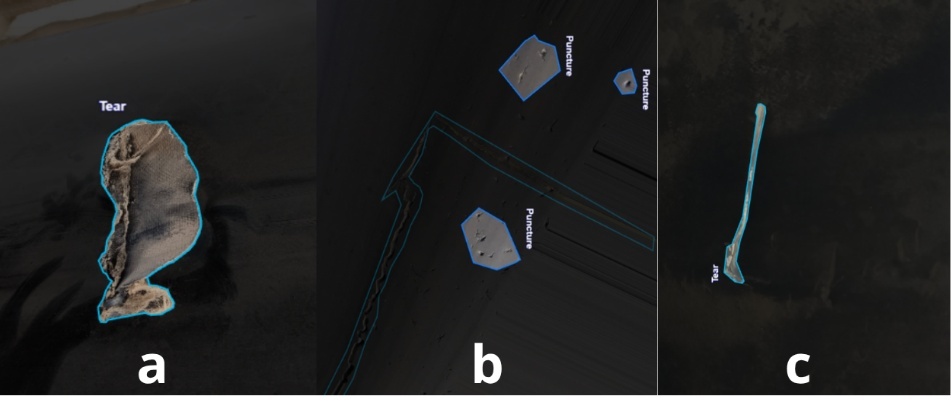}
    \caption{Corresponding annotated training samples showing ground-truth bounding box delineations across classes: (a) field inspection defects; (b) transformed public annotations; (c) sampled conveyor damage.}
    \label{fig:labeled_samples}
\end{figure}

Representative examples of the raw multi-source imagery and their corresponding annotated representations are depicted in Fig.~\ref{fig:raw_samples} and Fig.~\ref{fig:labeled_samples}.

To prevent overfitting and enhance generalization against industrial visual noise, an extensive augmentation pipeline was executed via Roboflow. Each raw sample generated three augmented variants via:
\begin{itemize}
    \item Spatial transformations: horizontal flip, vertical flip, $90^\circ$ clockwise and counter-clockwise rotations, and random rotations within $[-8^\circ, +8^\circ]$;
    \item Perspective shear: horizontal and vertical shearing of $\pm 15^\circ$;
    \item Photometric perturbations: grayscale conversion applied to 18\% of samples; hue variation of $\pm 15\%$, saturation adjustment of $\pm 33\%$, brightness modulation of $\pm 25\%$, and exposure shift of $\pm 11\%$;
    \item Sensor noise modeling: Gaussian blur up to 3.4 pixels and salt-and-pepper noise up to 0.46\% of total image pixels.
\end{itemize}

The combined dataset contains 1,233 original images, expanding to a final corpus of \textbf{3,007 augmented images}. The corpus was partitioned into training (88\%, 2,661 images), validation (8\%, 231 images), and held-out test (4\%, 115 images) subsets.

\subsection{YOLOv11 Architecture and Mathematical Formulation}
We selected YOLOv11 nano (YOLOv11n)~\cite{jocher2024yolo11} as the foundational architecture. YOLOv11 introduces architectural refinements over previous generations, utilizing C3k2 building blocks, enhanced Spatial Pyramid Pooling Fast (SPPF), and an anchor-free task-aligned detection head.

The model optimizes a composite multi-task loss function comprising task-aligned classification loss $\mathcal{L}_{\text{cls}}$ (Varifocal / Binary Cross-Entropy), bounding box regression loss $\mathcal{L}_{\text{box}}$ (Complete Intersection over Union, CIoU), and Distribution Focal Loss $\mathcal{L}_{\text{dfl}}$:
\begin{equation}
    \mathcal{L}_{\text{total}} = \lambda_{\text{cls}} \mathcal{L}_{\text{cls}} + \lambda_{\text{box}} \mathcal{L}_{\text{box}} + \lambda_{\text{dfl}} \mathcal{L}_{\text{dfl}}
    \label{eq:loss}
\end{equation}
where $\lambda_{\text{cls}}$, $\lambda_{\text{box}}$, and $\lambda_{\text{dfl}}$ denote balancing hyper-parameters. The bounding box overlap is quantified via the Intersection over Union metric:
\begin{equation}
    IoU = \frac{\text{Area}(B_p \cap B_g)}{\text{Area}(B_p \cup B_g)}
    \label{eq:iou}
\end{equation}
where $B_p$ and $B_g$ represent the predicted and ground-truth bounding coordinates, respectively.

\subsection{Post-Training Quantization and Edge TPU Compilation}
The Google Coral Edge TPU executes neural network computations exclusively using 8-bit fixed-point arithmetic. Consequently, the floating-point weights and activation tensors of the PyTorch-trained YOLOv11 model must undergo full integer post-training quantization (PTQ)~\cite{prokscha2023efficient}. Quantization maps real-valued floating-point numbers $r \in [\alpha, \beta]$ to an 8-bit integer space $q \in [-128, 127]$ through affine transformation:
\begin{equation}
    q = \text{clamp}\left( \left\lfloor \frac{r}{S} \right\rceil + Z, -128, 127 \right)
    \label{eq:quant}
\end{equation}
where $S$ is the scale factor and $Z$ is the zero-point offset:
\begin{equation}
    S = \frac{\beta - \alpha}{2^b - 1}, \quad Z = \text{round}\left( -\frac{\alpha}{S} \right) - 128
\end{equation}
with $b = 8$. A representative calibration dataset of 200 unlabelled training images was passed through the model to establish dynamic activation ranges.

Following TFLite export, the model was compiled using the Google Edge TPU Compiler (\texttt{edgetpu\_compiler -s}):
\begin{equation}
\begin{aligned}
    \text{Input:}  &\quad \texttt{model\_int8.tflite} \\
    &\quad \Big\downarrow\ \text{\scriptsize Edge TPU Compiler (\texttt{-s})} \\
    \text{Output:} &\quad \texttt{model\_int8\_edgetpu.tflite}
\end{aligned}
\label{eq:edgetpu_compile}
\end{equation}
The compiler mapped all 2D convolution, depthwise convolution, and activation operators onto the Edge TPU's internal systolic array, leaving only light host-side post-processing (such as non-maximum suppression, NMS) to the Raspberry Pi 5 ARM Cortex-A76 CPU.

\section{Experimental Results and Industrial Evaluation}
\label{sec:results}

\subsection{Training Performance and Convergence Analysis}
The YOLOv11n model was trained for 100 epochs on an NVIDIA A100 GPU (80 GB VRAM) utilizing the Ultralytics framework with AdamW optimizer, initial learning rate of $10^{-3}$, cosine annealing learning rate scheduler, batch size of 32, and an input resolution of $320 \times 320$ pixels.

\begin{figure}[!htbp]
    \centering
    \begin{subfigure}[b]{0.48\linewidth}
        \centering
        \includegraphics[width=\linewidth]{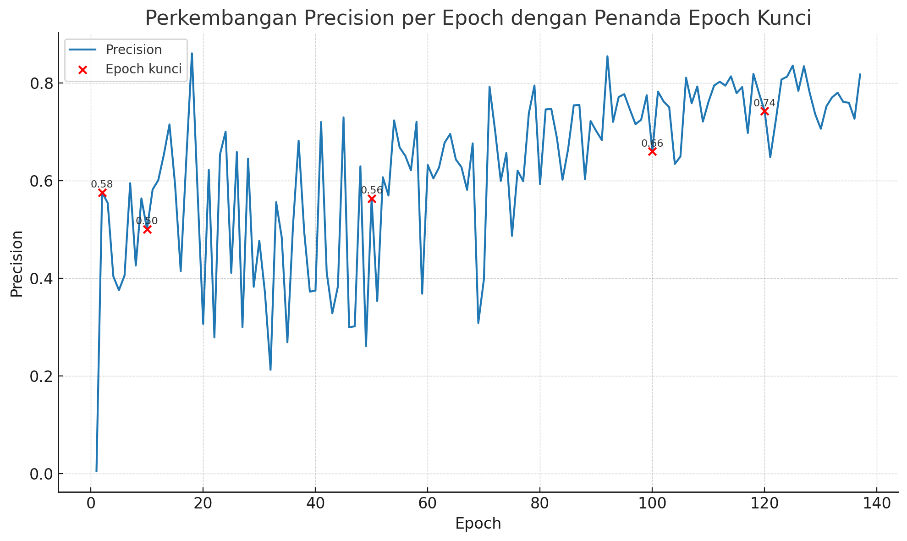}
        \caption{Precision vs. Epochs}
    \end{subfigure}
    \hfill
    \begin{subfigure}[b]{0.48\linewidth}
        \centering
        \includegraphics[width=\linewidth]{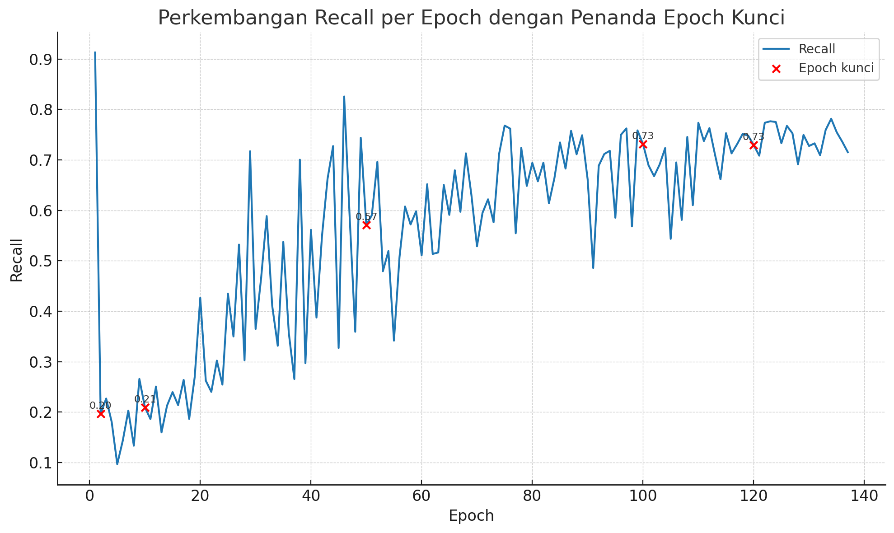}
        \caption{Recall vs. Epochs}
    \end{subfigure}
    \vspace{0.4em}
    \begin{subfigure}[b]{0.48\linewidth}
        \centering
        \includegraphics[width=\linewidth]{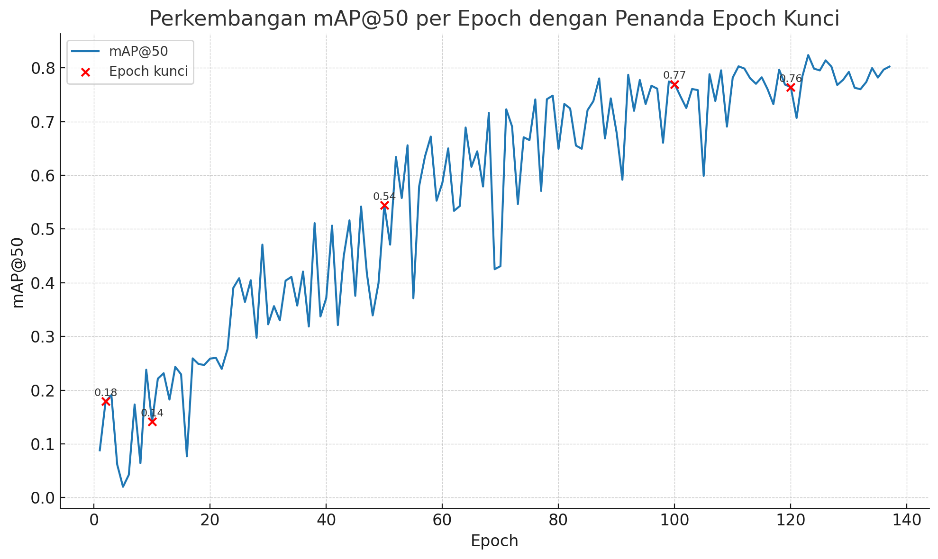}
        \caption{mAP@0.5 vs. Epochs}
    \end{subfigure}
    \hfill
    \begin{subfigure}[b]{0.48\linewidth}
        \centering
        \includegraphics[width=\linewidth]{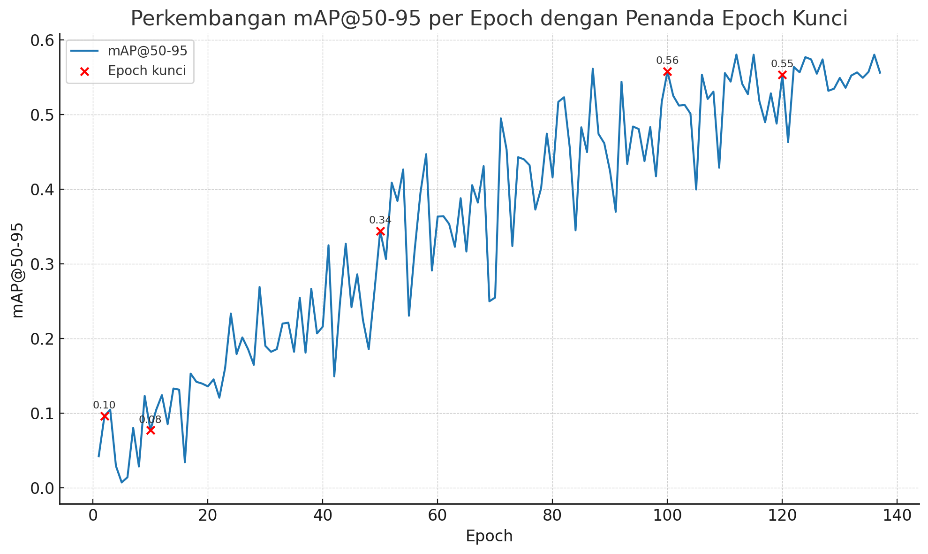}
        \caption{mAP@0.5:0.95 vs. Epochs}
    \end{subfigure}
    \caption{Training and validation performance progression of the YOLOv11 model across 100 epochs: (a) Precision converging to 0.742; (b) Recall converging to 0.730; (c) mAP@0.5 reaching 0.764; (d) mAP@0.5:0.95 reaching 0.553.}
    \label{fig:metrics}
\end{figure}

As shown in Fig.~\ref{fig:metrics}, all four primary validation metrics progressed smoothly without evidence of catastrophic divergence or severe overfitting~\cite{afaq2020significance}. The final quantitative evaluation on the validation corpus yielded:
\begin{itemize}
    \item \textbf{Precision ($P$):} 0.742 (74.2\%)
    \item \textbf{Recall ($R$):} 0.730 (73.0\%)
    \item \textbf{mAP@0.5:} 0.764 (76.4\%)
    \item \textbf{mAP@0.5:0.95:} 0.553 (55.3\%)
\end{itemize}

Analysis of individual defect classes revealed that the \textit{Patch Work} class achieved the highest detection accuracy, attributable to the pronounced textural contrast between rubber repair patches and worn belt vulcanization. The \textit{Tear} class achieved robust detection under varying ambient light, whereas small \textit{Puncture} and \textit{Hole} defects exhibited slightly lower recall due to clinker dust partially obscuring small cavity margins.

\begin{figure}[!htbp]
    \centering
    \includegraphics[width=\linewidth]{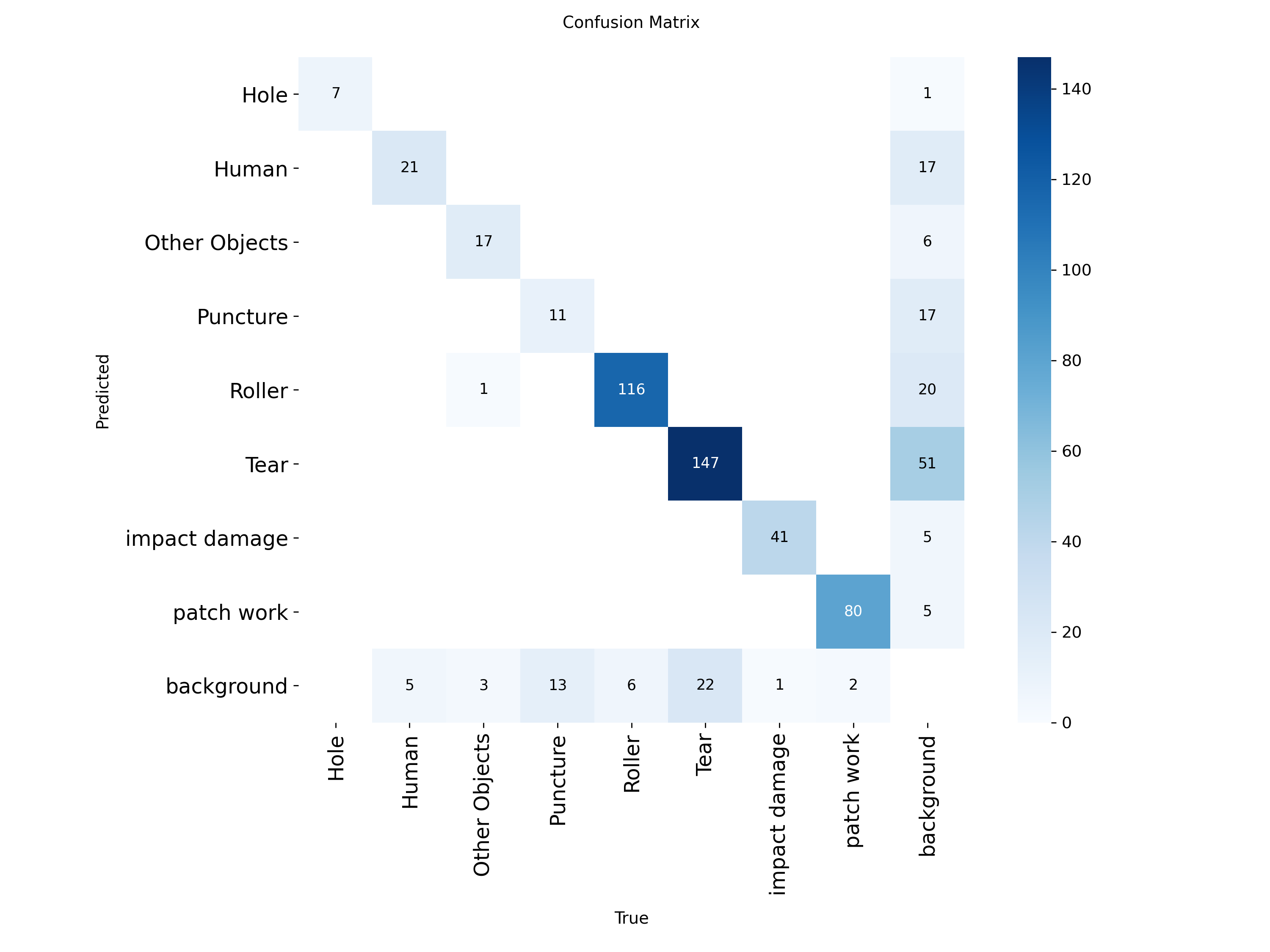}
    \caption{Normalized confusion matrix of the trained YOLOv11 model across the four defect categories and background, demonstrating robust true-positive localization with minimal inter-class confusion.}
    \label{fig:confusion_matrix}
\end{figure}

The normalized confusion matrix across all target categories is depicted in Fig.~\ref{fig:confusion_matrix}. The matrix confirms strong discriminatory performance on both \textit{Patch Work} (0.88) and \textit{Tear} (0.79) anomalies, while misclassifications between localized \textit{Punctures} and superficial dust patches represent the primary source of residual classification variance.

\subsection{Inference Speed and Hardware Acceleration Benchmarks}
To evaluate the impact of hardware acceleration and input resolution on edge inference performance, we benchmarked three distinct model deployment formats on the Raspberry Pi 5 platform:
\begin{enumerate}
    \item Standard PyTorch FP32 running on the host ARM CPU;
    \item Open Neural Network Exchange (ONNX) Runtime running on the host ARM CPU;
    \item Full integer INT8 TFLite running on the Google Coral Edge TPU coprocessor.
\end{enumerate}

Each configuration was evaluated across three input resolutions: $280 \times 280$, $320 \times 320$, and $640 \times 640$ pixels. The experimental findings are summarized in Table~\ref{tab:inference_speed}.

\begin{table*}[!t]
\centering
\small
\caption{Comparative Inference Throughput (Frames Per Second) on Raspberry Pi 5 Across Runtime Formats and Spatial Resolutions.}
\label{tab:inference_speed}
\begin{tabularx}{\textwidth}{lccccX}
\toprule
\textbf{Model Format} & \textbf{Device} & \textbf{280 px} & \textbf{320 px} & \textbf{640 px} & \textbf{Operational Behavior \& Observations} \\
\midrule
PyTorch (\texttt{.pt}) & RPi 5 CPU & $\sim$20 FPS & $\sim$10 FPS & 3--7 FPS & High CPU load; at 640 px bounding boxes appear with noticeable visual lag. \\
ONNX (CPU) & RPi 5 CPU & --- & --- & $\sim$20 FPS & Evaluated on CPU runtime; steady frame rate but elevated core temperatures. \\
\textbf{TFLite + Edge TPU} & \textbf{Coral USB} & --- & \textbf{30--40 FPS} & $\sim$10 FPS & \textbf{Highest throughput achieved at 320 px; stable real-time Patch Work and Tear detection.} \\
\bottomrule
\end{tabularx}
\end{table*}

The benchmark results in Table~\ref{tab:inference_speed} underscore the profound acceleration provided by the Edge TPU. At the native $320 \times 320$ resolution, the Edge TPU delivers a 3$\times$ to 4$\times$ speedup over the multi-core ARM CPU running PyTorch FP32, reaching \textbf{30--40 FPS}. At $640 \times 640$ pixels, inference throughput decreases to $\sim 10$ FPS due to memory transfer bandwidth limitations across the USB 3.0 interface and increased tensor dimensions. Consequently, $320 \times 320$ represents the optimal operational point balancing visual defect fidelity and real-time processing margin.

\subsection{In-Situ Industrial Field Deployment and Operational Stability}
The system was deployed on-site at an industrial cement manufacturing facility in West Java, Indonesia, to monitor an operational raw material clinker conveyor. The Raspberry Pi 5 ingested the RTSP sub-stream ($640 \times 360$ encoded in H.264 at 25 FPS) directly from the Hikvision IP camera over the dedicated Gigabit PoE link.

\begin{figure}[!htbp]
    \centering
    \begin{subfigure}[b]{0.48\linewidth}
        \centering
        \includegraphics[width=\linewidth]{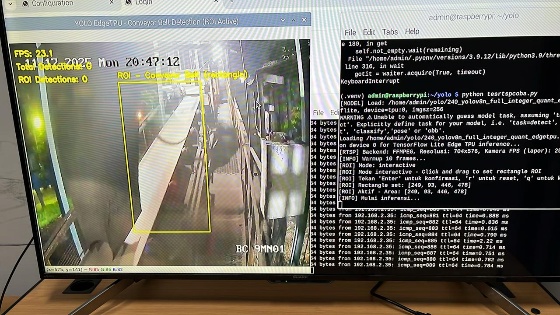}
        \caption{System interface}
    \end{subfigure}
    \hfill
    \begin{subfigure}[b]{0.48\linewidth}
        \centering
        \includegraphics[width=\linewidth]{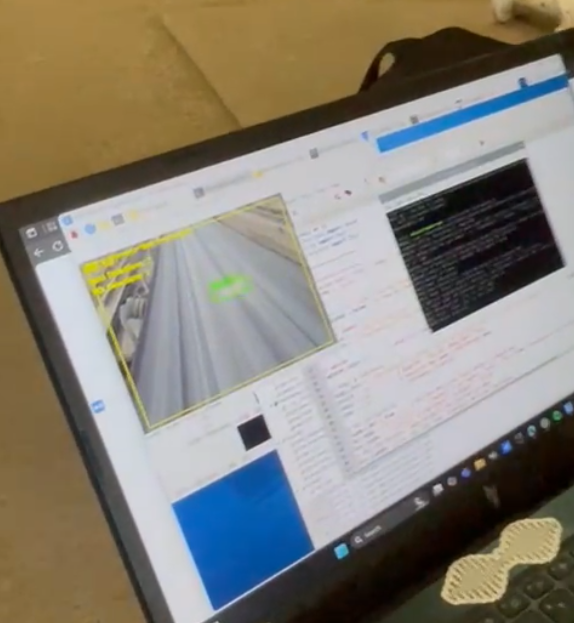}
        \caption{Live defect detection}
    \end{subfigure}
    \caption{Field deployment results: (a) system graphical user interface displaying real-time video feed and defect count; (b) positive real-time detection of a \textit{Patch Work} anomaly on the running industrial conveyor.}
    \label{fig:industrial_detection}
\end{figure}

As depicted in Fig.~\ref{fig:industrial_detection}, the system successfully localized surface defects in real time, drawing unambiguous bounding boxes with high confidence scores. To assess continuous operational viability, the system was subjected to prolonged continuous execution inside the sealed industrial enclosure box. The observed operational parameters are compiled in Table~\ref{tab:operational_stability}.

\begin{table*}[!t]
\centering
\small
\caption{System Operational and Environmental Parameters Recorded During Sustained Industrial Field Testing.}
\label{tab:operational_stability}
\begin{tabularx}{\textwidth}{lp{3.5cm}X}
\toprule
\textbf{Parameter} & \textbf{Measured Value} & \textbf{Operational Context \& Stability Status} \\
\midrule
Maximum Edge TPU FPS & 30--40 FPS & Measured using RTSP sub-stream at $320\times 320$ px; smooth defect tracking. \\
Minimum PyTorch FPS & 3--7 FPS & Host CPU at $640\times 640$ px; inadequate for fast conveyor speeds. \\
ONNX CPU Frame Rate & $\sim$20 FPS & General benchmark baseline on host CPU. \\
End-to-End Latency & $\mathbf{70\text{--}90\text{ ms}}$ & Cumulative latency from camera sensor exposure, network transport, preprocessing, TPU inference, to display render. \\
Raspberry Pi 5 Core Temp. & $55 \pm 5^\circ\text{C}$ & Thermally stable within enclosed metallic panel without active refrigeration. \\
Google Coral USB Temp. & $\sim 65^\circ\text{C}$ & Consistent thermal equilibrium under continuous 100\% TPU utilization. \\
RTSP Network Stream & Stable & Zero frame loss or memory leakage observed during prolonged testing. \\
\bottomrule
\end{tabularx}
\end{table*}

The system sustained an average end-to-end latency between 70 ms and 90 ms. Given an industrial conveyor speed of approximately 2.5 m/s, a latency of 80 ms corresponds to a linear belt displacement of only 0.20 m, ensuring that any detected defect is registered well before passing downstream pulleys. Thermal monitoring indicated that the Raspberry Pi 5 CPU maintained a steady temperature of $55 \pm 5^\circ\text{C}$ and the Coral USB coprocessor stabilized at $\sim 65^\circ\text{C}$, safely below thermal throttling thresholds.

\section{Discussion}
\label{sec:discussion}

\subsection{Architectural Trade-Offs in Edge AI Vision}
The experimental results demonstrate that deploying a lightweight, quantized deep learning model on edge hardware offers a compelling alternative to centralized server-based industrial inspection~\cite{ogden2020mdinference}. In conventional industrial architectures, streaming high-definition video from multiple conveyor lines to a central server incurs substantial network bandwidth overhead and vulnerability to plant-wide Wi-Fi or Ethernet dropouts. By performing inference directly at the edge node, bandwidth consumption is restricted to alert events and metadata, and decision latency is decoupled from network congestion.

Furthermore, post-training full integer quantization (INT8) reduced model parameter footprint by $\sim 75\%$ compared to single-precision FP32 with minimal degradation in detection accuracy ($\text{mAP@0.5} = 0.764$). The systolic tensor core execution on the Google Coral Edge TPU enabled a 4$\times$ increase in throughput compared to the quad-core Cortex-A76 host CPU while reducing host electrical power consumption~\cite{prokscha2023efficient}.

\subsection{Industrial Limitations and Environmental Sensitivity}
Despite the high detection rate achieved for major defects, several environmental challenges were documented during field trials~\cite{guo2022belt}:
\begin{enumerate}
    \item \textbf{Dust and Particulate Accumulation:} In the clinker handling area, airborne cement dust continuously deposits onto the conveyor belt and camera lens. Heavy dust layers reduce visual contrast between small punctures and surrounding rubber, occasionally generating false negatives. Periodic optical air-purge nozzles or automated wipers on the camera enclosure are recommended for long-term unattended operation.
    \item \textbf{Lighting Non-Uniformities:} Rapid transitions between harsh natural sunlight and shadowed indoor chutes can produce localized over-exposure. While the Hikvision camera's 120 dB WDR mitigated this significantly, supplementary LED ring illumination with polarized diffusers would further stabilize input image variance.
    \item \textbf{Small-Scale Defect Representation:} Fine hairline tears and micro-punctures remain challenging at $320 \times 320$ resolution due to spatial subsampling. Future iterations may adopt a tiled patch inference strategy (e.g., SAHI) executed selectively when belt vibrations exceed a threshold.
\end{enumerate}

\section{Conclusion and Future Work}
\label{sec:conclusion}

In this paper, we presented the design, implementation, and empirical validation of a real-time conveyor belt damage detection system based on YOLOv11 and the Google Coral Edge TPU, deployed on an embedded Raspberry Pi 5 platform in an active cement manufacturing facility. By compiling an INT8-quantized YOLOv11n model onto an Edge TPU coprocessor, the system achieves a throughput of 30--40 FPS at $320 \times 320$ resolution, an end-to-end latency of 70--90 ms, and a detection accuracy of 0.764 mAP@0.5 across four defect categories. In-situ industrial evaluations confirmed robust real-time defect localization, with stable thermal behavior inside a sealed protective enclosure and zero RTSP frame loss under harsh industrial conditions.

Future work will focus on: (1) evaluating end-to-end NMS-free paradigms (such as Ultralytics YOLO26~\cite{jocher2026yolo26}) to further eliminate host post-processing overhead; (2) integrating temporal tracking algorithms (such as ByteTrack) to eliminate duplicate defect alerts across consecutive frames; (2) interfacing the Raspberry Pi edge node directly with industrial Programmable Logic Controllers (PLCs) and Distributed Control Systems (DCS) via Modbus TCP and OPC UA to enable automated emergency conveyor stop triggers; and (3) deploying multi-spectral or thermal imaging modalities to detect subsurface cord degradation before external tears manifest.

\section*{Acknowledgments}
The authors express their sincere gratitude to the engineering and maintenance management team at the industrial cement manufacturing facility in West Java, Indonesia, for providing industrial facility access, technical collaboration, and operational conveyor data during the field research period. The authors also acknowledge the Department of Electrical Engineering at Universitas Jenderal Soedirman for providing academic guidance and laboratory support throughout this investigation.

\IfFileExists{main.bbl}{
}{\bibliographystyle{plain}\bibliography{references}}

\end{document}